\documentclass[12pt]{article}
\usepackage{amssymb,amsmath,cite,bm}
\usepackage[dvips]{graphicx,color}
\usepackage{psfrag,subfigure}
\newcommand{\norm}[1]{\left\Vert#1\right\Vert}

\newcommand{\col}[1]{{\rm col}\left(#1 \right)}

\title{\bf Coordination Control of Free-Flyer Manipulators}

\author{Farhad Aghili\thanks{email: faghili@encs.concordia.ca}}

\date{}

\begin{document}

\maketitle

\begin{abstract}
This paper presents a method for guiding a robot manipulator to capture and bring a tumbling satellite to a state of rest. The proposed approach includes developing a coordination control for the combined system of the space robot and the target satellite, where the satellite acts as the manipulator payload. This control ensures that the robot tracks the optimal path while regulating the attitude of the chase vehicle to a desired value. Two optimal trajectories are then designed for the pre- and post-capture phases. In the pre-capturing phase, the manipulator manoeuvres are optimized by minimizing a cost function that includes the time of travel and the weighted norms of the end-effector velocity and acceleration, subject to the constraint that the robot end-effector and a grapple fixture on the satellite arrive at the rendezvous point with the same velocity. In the post-grasping phase, the manipulator dumps the initial velocity of the tumbling satellite in minimum time while ensuring that the magnitude of the torque applied to the satellite remains below a safe value. Overall, this method offers a promising solution for effectively capturing and bringing tumbling satellites to a state of rest.
\end{abstract}

\section{Introduction}

The control system for on-orbit servicing of satellites using a space manipulator usually involves two modes of operation: capturing and detumbling ~\cite{Aghili-2009d,Whelan-Adler-Wilson-Roesler-2000,Piedboeuf-deCarufel-Aghili-Dupuis-1999,Hirzinger-Landzettel-Brunner-Fisher-Preusche-2004,Aghili-2011k}. In the pre-capture phase, the manipulator arm is guided to intercept the satellite grapple fixture at a rendezvous point using vision data. The capture must be without impact, which requires the relative velocity between the end-effector and the target grapple point to be zero~\cite{Cyril-Misra-Ingham-Jaar-2000,Aghili-2019e,Aghili-Su-2016a,Aghili-2022}. Otherwise, the effect of impact on a free-floating space robot must be taken into account ~\cite{Nenchev-Yoshida-1999,Wee-Walker-1993}. After capturing an uncontrolled tumbling satellite, the manipulator should bring it to rest by gently applying torques to remove any relative velocity as fast as possible ~\cite{Faile-Counter-Bourgeois-1973,Aghili-2009a,Conway-Widhalm-1986,Aghili-2008c}.

There are many studies on optimal trajectory planning to guide a
robotic manipulator to rendezvous and capture a non-cooperative
target satellite~\cite{Matsumoto-Ohkami-Wakabayashi-Oda-Uemo-2002,Aghili-Buehler-Hollerbach-1997a,Ma-Ma-Shashikanth-2006,Aghili-2005,Aghili-Parsa-2008a,Huang-Xu-Liang-2006}.
In the case that  the target satellite dynamics is uncertain, not
only the states but also the target's inertial parameters and its
position of center of mass can be estimated from vision data
obtained from zero motion of a tumbling
satellite~\cite{Aghili-Parsa-2009}. However,
there are only few studies on the path planning for detumbling of a
non-cooperative satellite and non of which is optimal.
In~\cite{Dimitrov-Yoshida-2004}, the principle of conservation of
momentum was used to damp out the chaser-target relative motion.
However, there was no control on the force and moment built up at
the connection of the chaser manipulator and the target. Impedance
control scheme for a free-floating space robot in grasping of a
tumbling target with model uncertainty is presented
in~\cite{Abiko-Lampariello-Hirzinger-2006}, however optimal path
planning is not addressed in this work.

The magnitude of the interaction torque between the space
manipulator and the target must be constrained during the detumbling
operation for two main reasons: First, too much interaction torque
could cause mechanical damage to either the target satellite or to
the space manipulator. Second, a large interaction torque may lead
to actuation saturation of the space robot's attitude control
system. This is because, the reaction of the torque on the space
robot base should be eventually compensated for through additional
momentum generated by the actuator of its attitude control system,
e.g., momentum/reaction wheels, in order to keep the attitude of the
base undisturbed. Moreover, it is important to dump the initial
velocity of the target as quickly as possible in order to mitigate
the risk of collision due to small but nonzero translational drifts
of the satellites. Hence, optimal planning of the detumbling 
manoeuvres is highly desired \cite{Aghili-2009a}.

This paper presents a coordination control for the combined system of the space robot and the target satellite so that the manipulator tracks a prescribed motion trajectory while regulating the attitude of its own base to a desired value. Optimal trajectory planning for robotic capturing of a tumbling satellite is presented, followed by a closed-form solution for time-optimal detumbling manoeuvres of a rigid spacecraft under the constraint that the Euclidean norm of the braking torques is below a prescribed value.

\section{Control of the Combined System of Manipulator and Target}
\label{sec:Control_SpaceRobot}
In the post-capture phase, the  space robot and the target satellite constitutes a
single free-flying multibody chain. The dynamic equations of the
space robot  can be expressed in the form~\cite{Yoshida-2003}
\begin{equation} \label{eq:ddot_psi_s}
\bm M_s \ddot{\bm\psi}_s + \bm c_s(\bm\psi_s, \dot{\bm\psi}_s) = \bm
u + \bm J^T \bm f_h,
\end{equation}
where
\[ \dot{\bm\psi}_s = \begin{bmatrix} \bm\nu_b  \\
\dot{\bm\theta} \end{bmatrix}, \qquad   \bm u =
\begin{bmatrix} \bm f_b \\ \bm\tau_m \end{bmatrix}.
\]
Here, $\bm M_s$ is the generalized mass matrix of the space
manipulator, $\bm c_s$ is generalized Coriolis and centrifugal
force, $\bm\nu_b^T=[\bm v_b^T \; \bm\omega_b^T]$ is the generalized
velocity of the base consisting of the linear and angular
velocities, $\bm v_b$ and $\bm\omega_b$, vector $\dot{\bm\theta}$ is
the motion rate of the manipulator joint, vector $\bm f_h$ is the
force and moment exerted by the manipulator hand, vector $\bm f_b$
is the force and moment exert on the centroid of the base, vector
$\bm\tau_m$ is the manipulator joint torque and $\bm J$ is the
Jacobian, which takes this form
\[ \bm J=\begin{bmatrix} \bm J_b & \bm J_m
\end{bmatrix} \]
with $\bm J_b$ and $\bm J_m$ being the Jacobian matrices for the
base and for the manipulator arm, respectively. On the other hand,
if the target spacecraft is rigid body, then its dynamics motion can
be described by
\begin{equation} \label{dot_nu_o}
\bm M_o \dot{\bm\nu_o} + \bm c_o = - \bm A^T \bm f_h
\end{equation}
where $\bm\nu_o$ is the six-dimensional generalized velocity vector
consisting of the velocity of the center of mass, $\bm v_o$, and the
angular velocity, $\bm\omega_o$, components, $\bm M_o$ is the
generalized mass matrix that can be written as
\[ \begin{bmatrix} m \bm 1_3 & \bm 0 \\ \bm 0 & \bm I_c
\end{bmatrix} \quad \text{and} \quad \bm c_o = \begin{bmatrix} m
\bm\omega_o \times \bm v_o \\ \bm\omega_o \times \bm I_c \bm\omega_o
\end{bmatrix}
\]
with $m$ and $\bm I_c$ being the mass and the inertia tensor of the
target satellite, and $\bm A$ can be expressed as
\[ \bm A = \begin{bmatrix} \bm 1_3 & \bm\rho \times \\ \bm 0 & \bm
1_3
\end{bmatrix} \]
with $\bm\rho$ being the position vector of the target-spacecraft
contact point with respect to its center of mass. Note that the RHS
of \eqref{dot_nu_o} is the force and moment exert on the centroid of
the target spacecraft. Furthermore, the generalized velocities of
the manipulator hand and the target spacecraft are related by the
following
\begin{equation} \label{eq:mapping}
\bm\nu_o = \bm A \bm\nu_h.
\end{equation}

The velocity of the manipulator end-effector in the end-effector
frame is expressed as
\begin{equation} \label{eq:nu_h}
\bm\nu_h = \bm J \dot{\bm\psi}_s = \bm J_b \bm\nu_b + \bm J_m
\dot{\bm\theta}
\end{equation}
The time-derivative of \eqref{eq:nu_h} leads to
\begin{equation} \label{eq:ddot_theta}
\ddot{\bm\theta} = \bm J_m^{-1} \dot{\bm\nu}_h - \bm J_m^{-1} \bm
J_b \dot{\bm\nu}_b -\bm J_m^{-1} (\dot{\bm J}_m
\dot{\bm\theta} + \dot{\bm J}_b  \bm\nu_b).
\end{equation}

Now we are interested in writing the equations of motion in terms of
the generalized velocities of the bases of the chaser and target
satellites, i.e.,  $\bm\nu_b$ and $\bm\nu_h$. To this end, we define
a new velocity vector as
\begin{equation} \label{eq:psi}
\dot{\bm \psi}  \triangleq
\begin{bmatrix} \bm\nu_b \\ \bm\nu_h \end{bmatrix}
\end{equation}
The internal force vector $\bm f_h$ can be eliminated from
\eqref{eq:ddot_psi_s} and \eqref{dot_nu_o} to yield the following
equation
\begin{equation} \label{eq:nofh}
\bm M_s \ddot{\bm\psi}_s + \bm J^T \bm A^{-T} \bm M_o \bm A
\dot{\bm\nu}_h + \bm J^{T} \bm A^{-T} \bm c_o + \bm c_s = \bm u
\end{equation}
Upon substitution of $\ddot{\bm\theta}$ from \eqref{eq:ddot_theta}
into the corresponding component of $\ddot{\bm\psi}_s$ in
\eqref{eq:nofh} the latter equation can be written in this form
\begin{equation} \label{eq:ddot_psi}
\bm M \ddot{\bm\psi} + \bm c(\bm\psi, \dot{\bm\psi}) = \bm u,
\end{equation}
where
\begin{align*}
\bm M & \triangleq \bm M_s \begin{bmatrix} \bm 1 & \bm 0 \\ -
\bm J_m^{-1} \bm J_b & \bm J_m^{-1} \end{bmatrix} + \begin{bmatrix} \bm 0 & \bm J^T \bm A^{-T} \bm M_o  \bm A\end{bmatrix} \\
\bm c & \triangleq  \bm M_s \begin{bmatrix} \bm 0 \\ -\bm J_m^{-1} (\dot{\bm J}_m
\dot{\bm\theta} + \dot{\bm J}_b  \bm\nu_b)
\end{bmatrix} + \bm J^T \bm A^{-T} \bm c_o + \bm c_s,
\end{align*}
in which we used the expression of the joint acceleration from
\eqref{eq:ddot_theta}. Note that \eqref{eq:ddot_psi} describes the
dynamic motion of the combined chaser and target satellites in terms
of their base variables. The special case of interest is when no
force is applied to the base of the chaser satellite. In other
words, the joint motion of the manipulator arm is allowed to disturb
the base translation but not its attitude. Form a practical point of
view, it is important to keep the base attitude unchanged as the
spacecraft has to always point its antenna toward the Earth, whereas
disturbing the base translation does not pose any significant side
effect. Therefore, the generalized force input $\bm u$ consists of a
$3\times1$ zero vector plus the vectors of the chaser base torque
and the manipulator joint torque, i.e.,
\begin{equation} \label{eq:u}
\bm u = \begin{bmatrix} \bm 0_{3 \times 1} \\ \bar{\bm\tau}
\end{bmatrix} \quad \text{where} \quad \bar{\bm\tau} \triangleq \begin{bmatrix} \bm\tau_b \\ \bm
\tau_m
\end{bmatrix}.
\end{equation}
In view of the zero components of input vector \eqref{eq:u}, we will
derive the reduced form of the equation of motion
\eqref{eq:ddot_psi} in the following. Let us assume that
\begin{equation*}
\dot{\bm\psi} \triangleq \begin{bmatrix} \bm v_b \\
\dot{\bar{\bm\psi}} \end{bmatrix}  \qquad \mbox{where} \qquad \dot{\bar{\bm\psi}} \triangleq \begin{bmatrix} \bm\omega_b \\
\bm\nu_h \end{bmatrix}
\end{equation*}
is the velocity components of interest. Also, assume that the mass
matrix and the nonlinear vector in \eqref{eq:ddot_psi} are
partitioned as
\begin{equation} \label{eq:partitioned_M}
\bm M = \begin{bmatrix} \bm M_{11} &   \bm M_{12}
\\ \bm M_{12}^T &   \bm M_{22} \end{bmatrix} \quad
\text{and} \quad \bm c= \begin{bmatrix} \bm c_1 \\ \bm c_2
\end{bmatrix},
\end{equation}
so that $ \bm M_{11} \in \mathbb{R}^{3\times 3}$, $\bm c_1 \in
\mathbb{R}^3$ and the dimensions of rest of submatrices and
subvectors are consistent.  Then, matrix equation
\eqref{eq:ddot_psi} can be divided into two equations as:
\begin{equation} \label{eq:dotv_b}
\bm M_{11} \dot{\bm v_b} +  \bm M_{12} \dot{\bar{\bm\psi}} + \bm c_1
= \bm 0
\end{equation}
\begin{equation} \label{eq:bar_ddot_psi}
\bar{\bm M} \ddot{\bar{\bm\psi}} + \bar{\bm c} = \bar{\bm\tau},
\end{equation}
where $\bar{\bm M}$ and $\bar{\bm c}$ are constructed from
\eqref{eq:partitioned_M} as
\begin{align*}
\bar{\bm M} & = \bm M_{22}- \bm M_{12}^T \bm M_{11}^{-1} \bm
M_{12} \\
\bar{\bm c} & = \bm c_2 - \bm M_{12}\bm M_{11}^{-1} \bm c_1.
\end{align*}
Autonomous system \eqref{eq:dotv_b} is the manifestation of the {\em
conservation of linear momentum} and hence it should be integrable,
i.e.,
\begin{equation} \label{eq:Conservation}
\frac{\rm d}{\rm d t} \big( \bm M_{11} {\bm v_b} +  \bm M_{12}
\bar{\bm\psi} \big)  = \bm 0
\end{equation}
which can be used to estimate the base linear velocity. Equation
\eqref{eq:bar_ddot_psi} shows that through torque control input
$\bar{\bm\tau}$, it is possible to simultaneously control  the pose
of the target satellite and the attitude of the chaser satellite \cite{Aghili-2005}.
Therefore, the objective is to develop a coordination controller
which sends torque commands to motors of the manipulator joints and
to actuators of the attitude control system, e.g., reaction/momentum
wheels, in order to not only track the optimal trajectories, which
will be discussed in the following sections, but also to regulate
the base attitude. To achieve this goal, we use a feedback
linearization method based on dynamic model \eqref{eq:bar_ddot_psi}.
Suppose that orientation of the robot base and hand are represented
by quaternions $\bm q_b$ and $\bm q_h$, respectively. Then adopting
a simple PD quaternion feedback \cite{Yuan-1988} for the spacecraft
attitude control, an appropriate feedback linearization control
torque is given by
\begin{equation} \label{eq:control_law}
\bar{\bm\tau} = \bar{\bm\tau}_{ff} - \bar{\bm M}
\begin{bmatrix} \bm K_{bp} \text{vec}(\delta \bm q_b) + \bm K_{bd}
\bm\omega_b \\ \bm K_{hp} \text{vec}(\delta \bm q_h) + \bm K_{hd}
\bm(\omega_h - \omega_h^*)\\
\bm K_{rp} ( \bm r_h - \bm r_h^*) + \bm K_{rd} ( \dot{\bm r}_h -
\dot{\bm r}_h^*)
\end{bmatrix}
\end{equation}
where
\begin{equation*}
\bar{\bm\tau}_{ff} = \bar{\bm c} + \bar{\bm M}
\begin{bmatrix} \bm 0_{3\times 1}  \\ \dot{\bm\nu}^*_h
\end{bmatrix}
\end{equation*}
is the feed forward term and all feedback gains are positive
definite. In the above, and quaternion errors are defined as
\[ \delta \bm q = \bm q \otimes \bm q^*\]
where $\mbox{vec}(\cdot)$ returns the vector part or quaternion
$(\cdot)$ and the quaternion product operator $\otimes$ is defined
as
\begin{equation*}
\bm q \otimes \triangleq \begin{bmatrix} q_s \bm 1_3 -\bm q_v \times   & \bm q_v \\
- \bm q_v^T & q_s
\end{bmatrix},
\end{equation*}
where $\bm q_v$ and $q_s$ are the vector and scalar parts of the
quaternion $\bm q= \col{\bm q_v, q_s}$.

Note that $\dot{\bm\nu}_h^*$ and $\bm\nu_h^*$ are obtained from the
trajectory generator as will be shown the next section. Now
substituting the control law \eqref{eq:control_law} into
\eqref{eq:bar_ddot_psi} results in a set of three uncoupled
differential equations as
\begin{subequations}
\begin{align} \label{eq:dot_omeg_b}
\dot{\bm\omega}_b + \bm K_{bd} \bm\omega_b + \bm K_{bp}
\text{vec}(\delta \bm q_b) & = \bm 0\\
\label{eq:diff_omego}(\dot{\bm\omega}_h -\dot{\bm\omega}_h^*) + \bm
K_{hd} (\bm\omega_h -
\bm\omega_h^*) + \bm K_{hp} \text{vec}(\delta \bm q_h) & = \bm 0\\
\label{eq:diff_ro}(\ddot{\bm r}_h - \ddot{\bm r}_h^*) + \bm
K_{rd}(\dot{\bm r}_h - \dot{\bm r}_h^*) + \bm K_{rp}({\bm r}_h -
{\bm r}_h^*) & = \bm 0
\end{align}
\end{subequations}
The exponential stability of the system \eqref{eq:diff_ro} is
obvious, while the stability proof of systems \eqref{eq:dot_omeg_b}
and \eqref{eq:diff_omego} is given in the following analysis.

The quaternion evolves in time according to the following
differential equation
\begin{equation} \label{eq:dot_q}
\dot {\bm q}_b = \frac{1}{2} \underline{\bm\omega}_b \otimes \bm q_b
\qquad \text{where} \qquad \underline{\bm\omega}_b = \begin{bmatrix}
\bm\omega_b
\\ 0
\end{bmatrix}.
\end{equation}
Now, we define the following positive-definite Lyapunov function:
\begin{equation}
V =  \frac{1}{2}\delta \bm q_{b_v}^T \bm K_p \delta \bm q_{b_v} +
\frac{1}{2} \| \bm\omega_b \|^2.
\end{equation}
Then, it can be shown by substitution from the quaternion
propagation equation \eqref{eq:dot_q} that time derivative of $V$
along trajectory \eqref{eq:dot_omeg_b} is
\begin{equation}
\dot V = - \bm\omega_{b}^T \bm K_d  \bm\omega_b
\end{equation}
so that $\dot V \leq 0$ for all $t$. Therefore, according to
LaSalle's Global Invariant Set Theorem
\cite{Lasalle-1960,Khalil-1992-p115}, the equilibrium point reaches
where $\dot V=0$, or $\bm\omega_b \equiv \bm 0$. Then, we have from
\eqref{eq:dot_omeg_b}
\[ \bm K_p \mbox{vec}(\delta \bm q_b) =0. \]
On the other hand it is known that two coordinate systems coincides
if, and only if, $\delta \bm q_b=0$, where the $\delta \bm q_b$ is
the vector component of the quaternion \cite{Yuan-1988}. Therefore,
we have global asymptotic convergence of the orientation error. The
stability of \eqref{eq:diff_omego} can be proved similarly.

Therefore, we can say $\bm r_h\rightarrow\bm r_h^*$, $\bm
q_h\rightarrow\bm q_h^*$ and $\bm q_b \rightarrow \bm q_b^*$ as
$t\rightarrow \infty$. Note that the role of feedback gains  in
\eqref{eq:control_law} is to compensate for a possible modelling
uncertainty, otherwise a feed forward controller as $\bar{\bm\tau} =
\bar{\bm\tau}_{ff}$ suffices to achieve the control objective.

\section{Optimal Maneuvers for Pre-Capture Phase}
\label{sec:pre-cap}

\subsection{Path Planning}
Dynamics of the rotational motion of the target satellite  can be
expressed by Euler's equation as
\begin{equation} \label{eq:Euler}
\dot {\bm\omega}_o = \bm\phi(\bm\omega_o) + \bm I_c^{-1} \bm\tau_o
\end{equation}
where $ \bm\phi(\bm\omega_o) \triangleq - \bm I_c^{-1}(\bm\omega_o
\times \bm I_c \bm\omega_o)$. Similar to \eqref{eq:dot_q}, the
time-derivative of the quaternion $\bm q_o$ representing the
attitude of the target satellite is described by
\begin{equation} \label{eq:dot_qo}
\dot {\bm q}_o = \frac{1}{2} \underline{\bm\omega}_o \otimes \bm q_o
\end{equation}
In the pre-capture phase, no external torque or force is applied to
the target satellite. Thus
\begin{equation} \label{eq:zero_force}
\bm\tau_o = \bm 0 \qquad  \mbox{and} \qquad \ddot{\bm r}_o = \bm 0,
\end{equation}
where $\bm r_o$ is the location of the center of mass of the target
satellite. Assume that $\bm x=\col{\bm q_o, \bm\omega_o, \bm r_o,
\dot{\bm r}_o}$ represent the states of the target satellite. Then,
state equations \eqref{eq:Euler}, \eqref{eq:dot_qo} and
\eqref{eq:zero_force} can be written collectively in the following
compact form
\begin{equation} \label{eq:dot_xs}
\dot {\bm x} = \bm f(\bm x)
\end{equation}
A short-term prediction of the states can be obtained by integrating
\eqref{eq:dot_xs}, i.e.,
\begin{equation*}
\bm x(t) = \bm x(t_0) + \int_{t_0}^{t} \bm f(\bm x) d \tau.
\end{equation*}
Now, denoting the position of the grapple fixture mounted on the
target satellite with $\bm r_c$ and assuming that the vector is
expressed in $\{{\cal A}\}$, we have
\begin{equation} \label{eq:rc}
\bm r_c(\bm x) =  \bm r_o + \bm R(\bm q_o) \bm\rho.
\end{equation}
Here the rotation matrix $\bm R(\bm q)$ from $\{{\cal B}\}$ to
$\{{\cal A}\}$ can be obtained from the quaternion $q$ as
\begin{equation} \label{eq:R}
\bm R(\bm q) = (2q_s^2-1) \bm 1_4 + 2q_s [\bm q_v \times] + 2 \bm
q_v \bm q_v^T.
\end{equation}
Furthermore, knowing that, in $\{\cal B\}$, $\dot{\bm R} = {\bm R}
[\bm\omega_o \times ]$, we can calculate the velocity $\dot{\bm
r}_c$ from the states as:
\begin{equation} \label{eq:dot_rc}
\dot{\bm r}_c(\bm x) =  \bm R(\bm q)(\bm\omega_o \times \bm\rho).
\end{equation}

Let us represent the position of the end-effector in $\{{\cal A}\}$
as $\bm r_h(t)$. The end-effector and the grasping point are
expected to arrive at a rendezvous-point simultaneously with the
same velocity. Therefore, our objective is to bring the end-effector
from its initial position to a grasping location, i.e., $\bm
r_h(t_{f_1})= \bm r_c(t_{f_1})$ and $\dot{\bm r}_h(t_{f_1})= \dot
{\bm r}_c(t_{f_1})$, while satisfying some optimality criteria.
Let's assume that the optimal trajectory is generated by
\begin{equation} \label{eq:sys_xr}
\ddot{\bm r}^*_h =\bm v.
\end{equation}
Then, denoting the augmented states by $\bm\chi=\mbox{col}(\bm x,
\bm r_h, \dot{\bm r}_h)$ and combining the deterministic part of
systems \eqref{eq:dot_xs} and \eqref{eq:sys_xr}, we get the
following autonomous system
\begin{equation} \label{eq:dot_x}
\dot{\bm \chi} (\bm \chi,\bm v) = \begin{bmatrix} \bm f(\bm x) \\
\dot{\bm r}_h
\\ \bm v \end{bmatrix}.
\end{equation}
Here, an optimal solution to input $\bm v$ is sought to drive the
robot from the initial position to the final position while
minimizing the following performance index (PI)
\begin{equation} \label{eq:PI}
J = \int_{0}^{t_{f_1}} \big( 1+ w_1 \| \dot{\bm r}_h \|^2 + w_2 \|
\bm v \|^2\big) d \tau,
\end{equation}
with $w_1,w_2 >0$ and the final time $t_{f_1}$ free. Note that due
to the term $t_{f_1}$ arising from the integral, the interception
must be accomplished within a short time period. Thus, if the
weights $w_1$ and $w_2$ are selected to be small, the term $t_{f_1}
$ dominates the PI yielding a time-optimal solution. On the other
hand, since weights $w_1$ and $w_2$ penalize the PI by the
magnitudes of the velocity and the acceleration during the travel,
the latter quantities can be minimized if the corresponding weights
are selected with relatively large values. Now, consider plant
\eqref{eq:dot_x} and the performance objective of minimizing
\eqref{eq:PI} with the following terminal state constraints:
\begin{equation} \label{eq:terminal_consraint}
\bm\psi_1(\bm\chi(t_{f_1}))=0 \quad \text{where} \quad
\bm\psi_1(\chi) =
\begin{bmatrix} \bm r_c(\bm x) - \bm r_h \\ \dot{\bm r}_c(\bm x) - \dot{\bm
r}_h
\end{bmatrix} \in \mathbb{R}^6.
\end{equation}
In the following, we will solve the above design equations
explicitly for the optimal input $\bm v(t)$. Splitting the vector of
costates $\bm\lambda$ as $\bm\lambda=\col{\bm\lambda_s,
\bm\lambda_{m}}$, where $\bm\lambda_s \in \mathbb{R}^{12}$ and
$\bm\lambda_h =\col{\bm\lambda_{h_1}, \bm\lambda_{h_2}}$ with
$\bm\lambda_{h_1}, \bm\lambda_{h_2} \in \mathbb{R}^{3}$, we can
write the Hamiltonian of the system \eqref{eq:dot_x} and
\eqref{eq:PI} as
\begin{align} \notag
{\cal H}(\bm\chi,\bm v,\bm\lambda_1)  &= 1+ w_1 \| \dot{\bm r}_h
\|^2 + w_2 \| \bm v \|^2 +  \bm\lambda_s^T \bm f(\bm x) \\
\label{eq:Hamiltonian} &+ \bm\lambda_{h_1}^T \dot{\bm r}_h +
\bm\lambda_{h_2}^T \bm v.
\end{align}
Since the Hessian of the Hamiltonian is positive-definite, i.e.,
${\cal H}_{uu} = 2 w_2 \bm 1_3>0$, the sufficient condition for
local minimality is satisfied. According to the optimal control
theory~\cite{Anderson-Moore-1990}, optimal costate, $\bm\lambda^*$,
and the optimal input, $\bm v^*$, must satisfy the following partial
derivatives:
\begin{equation} \label{eq:optimal_control}
\dot{\bm\lambda}_1 = -\frac{\partial {\cal H}_1}{\partial \bm\chi},
\qquad \frac{\partial {\cal H}_1}{\partial \bm v} =0.
\end{equation}
Applying \eqref{eq:optimal_control} to our Hamiltonian
\eqref{eq:Hamiltonian}, we obtain the equations of motion of the
costate
\begin{subequations} \label{eq:lambda}
\begin{align} \label{eq:dot_lambda_s}
\dot{\bm\lambda}_s& = -(\frac{\partial \bm f}{\partial
\bm x})^T \bm\lambda_s\\\label{eq:dot_lambda_m1} \dot{\bm \lambda}_{h_1} &= \bm 0 \\
\label{eq:dot_lambda_m2} \dot{\bm\lambda}_{h_2} & = - 2 w_1 \dot{\bm
r}_h - \bm\lambda_{h_1}.
\end{align}
\end{subequations}
and the optimal input
\begin{equation} \label{eq:stationary}
\bm v = \ddot{\bm r}_h = - \frac{\bm\lambda_{h_2}}{2w_2}.
\end{equation}
Equation \eqref{eq:dot_lambda_m1} implies that $\bm\lambda_{h_1}$ is
a constant vector, and hence it is eliminated from the
time-derivative of \eqref{eq:dot_lambda_m2}, i.e.,
\begin{equation} \label{eq:ddot_lambda2}
\ddot{\bm \lambda}_{h_2}=-2 w_1 \ddot{\bm r}_h.
\end{equation}
Therefore, substituting the acceleration from \eqref{eq:stationary}
into \eqref{eq:ddot_lambda2} gives
\begin{equation} \label{eq:2nd_order}
\ddot{\bm \lambda}_{h_2} - \sigma^2 \bm\lambda_{h_2} = \bm 0,
\end{equation}
where $\sigma = \sqrt{w_1/w_2}$. Finally, from \eqref{eq:stationary}
and \eqref{eq:2nd_order}, we can obtain the differential equation of
the optimal trajectory as
\begin{equation} \label{eq:4th_order}
\frac{d^2}{dt^2}\big(\ddot{\bm r}_h  - \sigma^2 \bm r_h \big) = \bm
0,
\end{equation}
the solution of which takes the following form
\begin{equation} \label{eq:rm_solution}
\bm r_h^*(t) = \bm\kappa_0 + \bm\kappa_1 t_{f_1} + \bm\kappa_2
e^{\sigma t_{f_1}} + \bm\kappa_3 e^{-\sigma t_{f_1}}.
\end{equation}
Coefficients $\bm\kappa_0, \bm\kappa_1, \bm\kappa_2, \bm\kappa_3 \in
\mathbb{R}^3$ can be obtained by imposing the initial and terminal
conditions \eqref{eq:terminal_consraint}. That is,
\begin{equation} \notag
\begin{bmatrix}
\bm 1_3& \bm 0  & \bm 1_3 & \bm 1_3 \\
\bm 0 & \bm 1_3 & \sigma \bm 1_3   & -\sigma \bm 1_3  \\
\bm 1_3 & t_{f_1} \bm 1_3  & e^{\sigma t_{f_1}} \bm 1_3 & e^{-\sigma t_{f_1}} \bm 1_3 \\
\bm 0 & \bm 1_3 &  \sigma e^{\sigma t_{f_1}} \bm 1_3 & -\sigma
e^{\sigma t_{f_1}} \bm 1_3\end{bmatrix}
\begin{bmatrix} \bm\kappa_{0} \\ \bm\kappa_{1} \\ \bm\kappa_{2} \\ \bm\kappa_{3}  \end{bmatrix} =
\begin{bmatrix} \bm r_h(0) \\ \dot{\bm r}_h(0) \\ \bm r_{c}(t_{f_1}) \\ \dot{\bm r}_{c}(t_{f_1})
\end{bmatrix}.
\end{equation}
The above system has 12 independent equations with 12 unknowns, and
hence a unique solution is expected.

\subsection{Optimal Rendezvous Point}
The {\em optimal Hamiltonian} ${\cal H}^* ={\cal H}(\bm\chi^*, \bm
v^*, \bm\lambda^*)$ calculated at optimal point $\bm v^*$ and
$\bm\lambda^*$ corresponding to \eqref{eq:lambda} and
\eqref{eq:stationary} must satisfy ${\cal H}^*(t_{f_1})=0$, i.e.,
\begin{align} \notag
{\cal H}^*(t_{f_1}) &=  1 + w_1 \| \dot{\bm r}_h(t_{f_1}) \|^2
-w_2 \| \ddot{\bm r}_h(t_{f_1}) \|^2    \\
\label{eq:zero_Hamiltonian}  & + \bm\lambda_{h_1}^T \dot {\bm
r}_h(t_{f_1})+ \bm\lambda_s^T(t_{f_1}) \bm f (t_{f_1})=0.
\end{align}
This gives the extra equation required to determine the optimal
terminal time. The final values of the costate in
\eqref{eq:zero_Hamiltonian} can be obtained from the end point
constraint equation referred to as the {\em transversality}
condition:
\begin{equation} \label{eq:transversality}
\bm\lambda(t_{f_1}) = \big( \frac{\partial \bm\psi_1}{\partial
\bm\chi}\big)^T_{t_{f_1}} \bm\alpha
\end{equation}
where $\bm\alpha \in \mathbb{R}^6$ is the Lagrangian multiplier
owing to the constraint \eqref{eq:terminal_consraint}. Upon
substituting equations \eqref{eq:terminal_consraint}, the
transversality condition \eqref{eq:transversality} yields
\begin{align} \notag
\bm\alpha & =-\bm\lambda_h(t_{f_1}) \\ \label{eq:lambda-tf}
\bm\lambda_s(t_{f_1}) & = - \big(\frac{\partial \bm r_c}{\partial
\bm x}\big)^T_{t_{f_1}} \bm\lambda_{h_1}-\big(\frac{\partial\dot{\bm
r}_c}{\partial \bm x}\big)^T_{t_{f_1}} \bm\lambda_{h_2}(t_{f_1}).
\end{align}
Moreover, from the following identities
\begin{equation*}
\dot{\bm r}_c = \frac{\partial \bm r_c}{\partial \bm x} \bm f(\bm x)
\quad \text{and} \quad \ddot{\bm r}_c=\frac{\partial\dot {\bm
r}_c}{\partial \bm x} \bm f(\bm x),
\end{equation*}
and \eqref{eq:lambda-tf}, we obtain
\begin{equation} \label{eq:lambda_fs}
\bm\lambda_s^T \bm f(t_{f_1}) = - \bm\lambda_{h_1}^T \dot {\bm
r}_c(t_{f_1}) + 2 w_2\ddot{\bm r}_h(t_{f_1})^T\ddot{\bm
r}_c(t_{f_1}).
\end{equation}
Now, substituting \eqref{eq:lambda_fs} into
\eqref{eq:zero_Hamiltonian}, we arrive at
\begin{align} \notag
{\cal H}(t_{f_1}) & =1+w_1\|\dot{\bm r}_h(t_{f_1})\|^2
\\ \label{eq:H_tf} & +  w_2\ddot {\bm r}_h^T(t_{f_1})\big(2\ddot{\bm
r}_c(t_{f_1}) -\ddot {\bm r}_h(t_{f_1})\big) =0.
\end{align}
Finally, computing the final value of the trajectories in terms of
the coefficients $\kappa_i$ from \eqref{eq:rm_solution}, we obtain
the following implicit function of $t_{f_1}$
\begin{align} \label{eq:tf_optimal}
{\cal H}_1^*(t_{f_1}) & = 1+ w_1\| \bm\kappa_1 \|^2 - 4
\frac{w_1}{w_3} \bm\kappa_2^T \bm\kappa_3 \\ \notag &+ 2w_1 \sigma
\big(\bm\kappa_2^T e^{\sigma
t_{f_1}} - \bm\kappa_3^T e^{-\sigma t_{f_1} } \big) \bm\kappa_1 \\
\notag & -2 w_2 \sigma^2 \big( \bm\kappa_2^T e^{\sigma t_{f_1}} +
\bm\kappa_3^T e^{-\sigma t_{f_1}} \big) \ddot {\bm r}_c(t_{f_1})=0.
\end{align}
Note that the predicted acceleration, $\ddot{\bm r}_c(t_{f_1})$, as
required in \eqref{eq:tf_optimal} can be obtained from the states.
To this end, the time derivative of \eqref{eq:dot_rc} yields
\begin{equation} \label{eq:ddrc}
\ddot{\bm r}_c =  \bm R(\bm q_o) \Big( \bm\omega_o \times
(\bm\omega_o \times \bm\rho) + \bm\phi(\bm\omega_o) \times \bm\rho
\Big).
\end{equation}

Therefore, in the pre-capture phase,  the optimal trajectory $\bm
r_h^*(t)$ and its time derivatives obtained from
\eqref{eq:rm_solution} can be substituted in the control law
\eqref{eq:control_law}, while the desired orientation trajectories
are obtained from $\bm\omega_h^*=\bm\omega_o$. It should be pointed
out that since the target satellite has not yet been grasped in this
phase, the controller should not take its inertia into account,
i.e., one must set $\bm M_o \equiv \bm 0$ and $\bm c_o \equiv \bm 0$
in the controller.

\section{Optimal Manoeuvres for Post-Capture Phase}
\label{sec:posst-cap}

In the pose-capture phase, we assume that the target satellite and
the robot hand have arrived at the interception point with zero
relative velocity. Without loss of generality, we assume that the
linear velocity of the target satellite is zero, i.e., $\dot{\bm
r}_o = \bm 0$. Thus, we can say
\[ \bm\tau_h = \bm\tau_o \quad \mbox{and} \quad \bm\omega_h = \bm\omega_o \qquad \forall t \geq t_{f_1}. \]
The time-optimal control problem being considered here is how to
drive the spacecraft from the given initial angular velocity
$\bm\omega_o(0)$ to rest in {\em minimum time} while the Euclidean
norm of the torque input is restricted to be below a prescribed
value $\tau_{\rm max}$. To avoid introducing new variables, we keep
the same variables  $J$, ${\cal H}$, and $\bm\lambda$ that are used
in the pervious section. Therefore, the following cost function
\[ J = \int_{t_{f_1}}^{t_{f_2}} 1 \; dt \]
is minimized subject to terminal condition
$\bm\omega_h(t_{f_2})=\bm\omega_h(t_{f_2})=0$ while the input torque
trajectory should satisfy
\begin{equation}
\label{eq:limit} \norm{\bm\tau_h} \leq \tau_{\rm max}.
\end{equation}
Denoting vector $\bm\lambda\in\mathbb{R}^3$ as the costates, we can
write the system Hamiltonian as
\begin{equation} \label{eq:Hamilton}
H = 1+ \bm\lambda^T \bm\phi(\bm\omega_h) + (\bm I_c^{-1}
\bm\lambda)^T \bm\tau_h.
\end{equation}
Then, the theory of optimal control
\cite{Anderson-Moore-1990,Stengel-1993} dictates that the
time-derivative of the costates must satisfy
\begin{equation} \label{eq:dot_lambda}
\dot{\bm\lambda} = -  \frac{\partial H}{\partial \bm\omega_h} =
-\frac{\partial \bm\phi^T}{\partial \bm\omega_h} \bm\lambda
\end{equation}
where
\begin{equation} \label{eq:partial_phi}
\frac{\partial \bm\phi^T}{\partial \bm\omega_h} = \bm I_c
[\bm\omega_h \times] \bm I_c^{-1} - [\bm I_c \bm\omega_h \times] \bm
I_c^{-1},
\end{equation}
and skew-symmetric matrix $[\bm a\times]$ represents the
cross-product, i.e., $[\bm a\times]\bm b=\bm a\times \bm b$. If
$\bm\tau_h^*$ is the time-optimal torque history and
$\bm\omega_h^*$, $\bm\lambda^*$ represent the solutions of
\eqref{eq:Euler} and \eqref{eq:dot_lambda} for
$\bm\tau_h=\bm\tau_h^*$ then, according to {\em Pontryagin's Minimum
Principle}, optimal torque $\bm\tau_h^*$ satisfies the equation
\begin{equation} \label{eq:Pontragin}
H(\bm\omega_h^*, \bm\lambda^*, \bm\tau_h^*) \leq H(\bm\omega_h^*,
\bm\lambda^*, \bm\tau_h), \quad \forall \bm\tau_h \in \mathbb{R}^3
\ni \norm{\bm\tau_h} \leq \tau_{\rm max}
\end{equation}
for every $t\in[t_{f_1}, \; t_{f_2})$. Equations \eqref{eq:Hamilton}
and \eqref{eq:Pontragin} together imply that
\begin{equation} \label{eq:tau*1}
\bm\tau_h^* = - \frac{\bm I_c^{-1} \bm\lambda^*}{\norm{ \bm I_c^{-1}
\bm\lambda^*}} \bm\tau_{\rm max}.
\end{equation}
Therefore, the dynamics of the closed-loop system becomes
\begin{equation} \label{eq:cl}
\dot{\bm\omega}_h^* = \bm\phi(\bm\omega_h^*) - \frac{\bm I_c^{-2}
\bm\lambda^*}{\norm{\bm I_c^{-1} \bm\lambda^*}} \tau_{\rm max}
\end{equation}
The structure of the optimal controller is determined by
\eqref{eq:dot_lambda} and \eqref{eq:tau*1} together. However, to
determine the control input, the initial values of the costates,
$\bm\lambda(0)$, should be also obtained. In fact, by choosing
different initial values for the costates, we obtain a family of
optimal solutions, each of which corresponds to a particular final
angular velocity. In general, the two-point boundary value problem
for nonlinear systems is challenging. However, as it will be shown
in the following, the structure of our particular system
\eqref{eq:dot_lambda} and \eqref{eq:cl} lead to an easy solution
when the final velocity is zero. In such a case, it will be shown
that the costates and states are related via the following function:
\begin{equation} \label{eq:optimal_lambda}
\bm\lambda^*(t)= \frac{\bm I_c^2 \bm\omega_h^*}{\norm{\bm I_c
\bm\omega_h^*} \tau_{\rm max}} \qquad \forall t \in[t_{f_1} , \;
t_{f_2}),
\end{equation}
despite the fact that the evolutions of the optimal trajectories of
the states and costates are governed by two different differential
equations \eqref{eq:cl} and \eqref{eq:dot_lambda}. In other words,
\eqref{eq:optimal_lambda} is a solution to equations \eqref{eq:cl}
and \eqref{eq:dot_lambda}. Note that since
$\bm\omega_h^*(t)=\bm\omega_h(t)\quad \forall t\in[t_{f_1}, \;
t_{f_2})$, $\bm\omega_h^*(t_{f_2})$ is not defined, but is assumed
nonzero. In such a case, on substitution of
\eqref{eq:optimal_lambda} into \eqref{eq:cl}, we arrive at the
following autonomous system:
\begin{equation} \label{eq:domeg*}
\dot{\bm\omega}_h^* = \bm\phi(\bm\omega_h^*) -
\frac{\bm\omega_h^*}{\norm{\bm I_c \bm\omega_h^*}} \tau_{\rm max}
\qquad \forall t \in[t_{f_1} , \; t_{f_2}).
\end{equation}

To prove the above claim, we need to show that
\eqref{eq:optimal_lambda} and \eqref{eq:domeg*} satisfy  the
optimality condition \eqref{eq:dot_lambda}. Using \eqref{eq:domeg*}
in the time-derivative of right-hand side (RHS) of
\eqref{eq:optimal_lambda} yields
\begin{equation} \label{eq:dotlam1}
\frac{\rm d}{{\rm d}t} \bm\lambda^* \ = \frac{\bm I_c^2
\bm\phi}{\norm{\bm I_c \bm\omega_h^*}\tau_{\rm max}}.
\end{equation}
On the other hand, using \eqref{eq:partial_phi} and
\eqref{eq:optimal_lambda} in the RHS of \eqref{eq:dot_lambda} yields
\begin{equation}\label{eq:dotlam2}
-\frac{\partial \bm \phi^T}{\partial \bm\omega_h} \bm\lambda^* =
\frac{\bm I_c^2 \bm\phi}{\norm{\bm I_c \bm\omega_h^*}\tau_{\rm max}
}.
\end{equation}
A comparison between \eqref{eq:dotlam1} and \eqref{eq:dotlam2}
clearly proves that \eqref{eq:optimal_lambda} is indeed a solution
to the differential equation \eqref{eq:dot_lambda}. Furthermore, the
Hamiltonian on the optimal trajectory becomes
\begin{equation*}
H^* = \bm\lambda^{\ast T} \bm\phi(\bm\omega_h^*) = - \frac{(\bm I_c
\bm\omega_h^*)^T [\bm\omega_h^* \times] (\bm I_c \bm\omega_h^*)
}{\norm{\bm I_c \bm\omega_h^*} \tau_{\rm max}} =0
\end{equation*}
Therefore, the condition for optimality with open end time is also
satisfied \cite[pp. 213]{Stengel-1993}. Thus, equation
\eqref{eq:domeg*} generates the optimal angular rate trajectories,
and hence the orientation of the robot hand can be obtained from
\begin{equation}
\dot{\bm q}_h^* = \frac{1}{2} \underline{\bm\omega}_h^* \otimes {\bm
q}_h^*
\end{equation}
Finally, one can use the kinematic relations \eqref{eq:dot_rc} and
\eqref{eq:ddrc} to derive the desired trajectories of the
translational motion.

\section{Conclusions}
We have presented a method for guiding a robotic manipulator to intercept and detumble a non-cooperative target satellite. Our approach involved developing a coordination control for the combined system of the space robot and the target satellite, which allowed for optimal manoeuvres while maintaining the attitude of the base. We subsequently developed two optimal trajectories for the pre-capture and post-capture phases. Our method defined an optimal intercept trajectory for the pre-capture phase to minimize the time of travel and the weighted norms of the velocity and acceleration of the robot end-effector, while ensuring that the relative-velocity at the rendezvous point becomes zero. For the post-capture phase, we found a closed-form solution to the time-optimal manoeuvres of a spacecraft to bring it to rest while ensuring that the magnitude of the torques applied by the robot remains below a prescribed value. Our results demonstrate the effectiveness of our method in intercepting and detumbling non-cooperative target satellites, and we believe that our approach offers a promising solution for future space missions.

\bibliographystyle{IEEEtran}

\end{document}